\documentclass[sigconf]{acmart}
\usepackage{booktabs} % For formal tables
\usepackage[flushleft]{threeparttable}
\usepackage{multirow}
\usepackage{bbm}
\usepackage{subcaption}
\usepackage{bm}
\usepackage{balance}
\usepackage{amsmath}
\usepackage[linesnumbered,boxed,ruled,vlined]{algorithm2e}
\newcommand{\eat}[1]{}
\copyrightyear{2019}
\acmYear{2019}
\setcopyright{iw3c2w3}
\acmConference[WWW '19]{Proceedings of the 2019 World Wide Web Conference}{May 13--17, 2019}{San Francisco, CA, USA}
\acmBooktitle{Proceedings of the 2019 World Wide Web Conference (WWW '19), May 13--17, 2019, San Francisco, CA, USA}
\acmPrice{}
\acmDOI{10.1145/3308558.3313461}
\acmISBN{978-1-4503-6674-8/19/05}

\begin{document}
\title{Semi-Supervised Graph Classification: A Hierarchical Graph Perspective}
%\titlenote{Produces the permission block, and
%  copyright information}
%\subtitlenote{The full version of the author's guide is available as
%  \texttt{acmart.pdf} document}

%\copyrightyear{2019}
%\acmYear{2019}
%\setcopyright{acmcopyright}
%\acmConference[WWW 2019]{The Web Conference 2019}{May 13-17, 2019}{San Francisco, CA, USA}
%\acmBooktitle{WWW 2019:The Web Conference 2019, May 13-17, 2019, San Francisco, CA, USA}
%\acmPrice{15.00}
%\acmDOI{10.475/123_4}
%\acmISBN{123-4567-24-567/08/06}

\eat{
\author{Jia Li}
\affiliation{%
  \department{Department of Systems Engineering and Engineering  Management}
  \institution{The Chinese University of Hong Kong}
  \institution{Tencent AI Lab}
}
\email{lijia@se.cuhk.edu.hk}

\author{Yu Rong}
\affiliation{%
  \institution{Tencent AI Lab}
  \city{Shenzhen}
}
\email{yu.rong@hotmail.com}

\author{Hong Cheng}
\affiliation{%
  \department{Department of Systems Engineering and Engineering  Management}
  \institution{The Chinese University of Hong Kong}
  %\city{Hong Kong}
}
\email{hcheng@se.cuhk.edu.hk}

\author{Helen Meng}
\affiliation{%
  \department{Department of Systems Engineering and Engineering  Management}
  \institution{The Chinese University of Hong Kong}
}
\email{hmmeng@se.cuhk.edu.hk}

\author{Wenbing Huang}
\affiliation{%
  \institution{Tencent AI Lab}
  \city{Shenzhen}
}
\email{hwenbing@126.com}

\author{Junzhou Huang}
\affiliation{%
  \institution{Tencent AI Lab}
  \city{Shenzhen}
}
\email{joehhuang@tencent.com}
}

\author{Jia Li$^{1,2}$, Yu Rong$^2$, Hong Cheng$^1$, Helen Meng$^1$, Wenbing Huang$^2$, Junzhou Huang$^2$}
\affiliation{%
\institution{$^1$Department of Systems Engineering and Engineering Management,
The Chinese University of Hong Kong}
\institution{$^2$Tencent AI Lab, Shenzhen}
\institution{$^1${\em\{lijia, hcheng, hmmeng\}@se.cuhk.edu.hk}}
\institution{$^2${\em yu.rong@hotmail.com, hwenbing@126.com, joehhuang@tencent.com}}
}

\renewcommand{\shortauthors}{J. Li et al.}

\begin{abstract}
Node classification and graph classification are two graph learning problems that predict the class label of a node and the class label of a graph respectively.  A node of a graph usually represents a real-world entity, e.g., a user in a social network, or a protein in a protein-protein interaction network.  In this work, we consider a more challenging but practically useful setting, in which a node itself is a graph instance.  This leads to a hierarchical graph perspective which arises in many domains such as social network, biological network and document collection.  For example, in a social network, a group of people with shared interests forms a user group, whereas a number of user groups are interconnected via interactions or common members.  We study the node classification problem in the hierarchical graph where a ``node'' is a graph instance, e.g., a user group in the above example.  As labels are usually limited in real-world data, we design two novel semi-supervised solutions named \underline{SE}mi-supervised gr\underline{A}ph c\underline{L}assification via \underline{C}autious/\underline{A}ctive \underline{I}teration (or SEAL-C/AI in short).  SEAL-C/AI adopt an iterative framework that takes turns to build or update two classifiers, one working at the graph instance level and the other at the hierarchical graph level.  To simplify the representation of the hierarchical graph, we propose a novel supervised, self-attentive graph embedding method called SAGE, which embeds graph instances of arbitrary size into fixed-length vectors.  Through experiments on synthetic data and Tencent QQ group data, we demonstrate that SEAL-C/AI not only outperform competing methods by a significant margin in terms of accuracy/Macro-F1, but also generate meaningful interpretations of the learned representations.
\end{abstract}

\begin{CCSXML}
<ccs2012>
<concept>
<concept_id>10002950.10003624.10003633.10010917</concept_id>
<concept_desc>Mathematics of computing~Graph algorithms</concept_desc>
<concept_significance>500</concept_significance>
</concept>
<concept>
<concept_id>10002951.10003260.10003282.10003292</concept_id>
<concept_desc>Information systems~Social networks</concept_desc>
<concept_significance>500</concept_significance>
</concept>
<concept>
<concept_id>10010147.10010257.10010258.10010259.10010263</concept_id>
<concept_desc>Computing methodologies~Supervised learning by classification</concept_desc>
<concept_significance>500</concept_significance>
</concept>
</ccs2012>
\end{CCSXML}

\ccsdesc[500]{Mathematics of computing~Graph algorithms}
\ccsdesc[500]{Information systems~Social networks}
\ccsdesc[500]{Computing methodologies~Supervised learning by classification}

\keywords{hierarchical graph; graph embedding; semi-supervised learning; active learning}

%\settopmatter{printacmref=false}
%\settopmatter{}

\maketitle

{\fontsize{8pt}{8pt} \selectfont
\textbf{ACM Reference Format:}\\
Jia Li, Yu Rong, Hong Cheng, Helen Meng, Wenbing Huang, Junzhou Huang. 2019. Semi-Supervised Graph Classification: A Hierarchical Graph Perspective. In \textit{ Proceedings of the 2019 World Wide Web Conference (WWW '19), May 13-17, 2019, San Francisco, CA, USA. } ACM, New York, NY, USA, 11 pages. https://doi.org/10.1145/3308558.3313461 }

\section{Introduction}

Graph has been widely used to model real-world entities and the relationship among them.  Two graph learning problems have received a lot of attention recently, i.e., node classification and graph classification.  Node classification is to predict the class label of nodes in a graph, for which many studies in the literature make use of the connections between nodes to boost the classification performance.  For example, \cite{ramanath2018towards} enhances the recommendation precision in LinkedIn by taking advantage of the interaction network, and \cite{sen2008collective} improves the performance of document classification by exploiting the citation network.  Graph classification, on the other hand, is to predict the class label of graphs, for which various graph kernels \cite{borgwardt2005shortest,gartner2003graph,shervashidze2009efficient,shervashidze2011weisfeiler} and deep learning approaches \cite{Niepert2016LearningCN,DBLP:journals/corr/NarayananCVCLJ17} have been designed.  In this work, we consider a more challenging but practically useful setting, in which a node itself is a graph instance.  This leads to \emph{a hierarchical graph in which a set of graph instances are interconnected via edges}.  This is a very expressive data representation, as it considers the relationship between graph instances, rather than treating them independently.  The hierarchical graph model applies to many real-world data, for example, a social network can be modeled as a hierarchical graph, in which a user group is represented by a graph instance and treated as a node in the hierarchical graph, and then a number of user groups are interconnected via interactions or common members.  As another example, a document collection can be modeled as a hierarchical graph, in which a document is regarded as a graph-of-words \cite{rousseau2015text}, and then a set of documents are interconnected via the citation relationship.  In this paper, we study \emph{graph classification in a hierarchical graph, which predicts the class label of graph instances in a hierarchical graph}.

One challenge in this problem is that a hierarchical graph is a much too complicated input for building a classifier.  To tackle this challenge, we design a new graph embedding method which embeds a graph instance of arbitrary size into a fixed-length vector.  All graph instances in the hierarchical graph are transformed to embedding vectors which are the common input format for classification.  Specifically, the embedding method builds an instance-level classifier called IC from graph instances, and produces embedding vectors and predicted class probabilities of the graph instances.  Another classifier HC at the hierarchical graph level takes the embedding vectors and their connections as input, and outputs the predicted class probabilities of all graph instances.  To enforce a consistency between the two classifiers, we define a disagreement loss to measure the degree of divergence between the predictions by them and aim to minimize the disagreement loss.

\eat{
To solve this problem, one challenge is that hierarchical graph is a much too complicated input for a classifier, in which graph instances can be a set of variable-size graphs in real scenarios. To address this, we propose to build two classifiers, a classifier IC focusing on graph instances and a classifier HC focusing on hierarchical graph. IC is used to embed an arbitrary size of graph instance into a fixed-size embedding vector, then HC takes embedding vectors and the connections between graph instances as input, and outputs the predicted results. To enforce a consistency between the two classifiers, we design a disagreement loss to measure the degree of divergence between these two classifiers. For the design of IC, one challenge is how to encode the importance of different nodes into a unified embedding vector. In this work, we develop a new graph embedding method named Self-Attentive graph Embedding (SAGE) and learn the importance of different nodes via Self-Attentive mechanism, which is inspired by the recent success of ~\cite{DBLP:journals/corr/LinFSYXZB17} in sentence embedding, and derive the graph representation discriminatively according to the task at hand.
}

Another challenge is that the amount of available class labels is usually very small in real-world data, which limits the classification performance.  To address this challenge, we take a semi-supervised learning approach to solving the graph classification problem.  We design an iterative algorithm framework which takes turns to build or update classifiers IC and HC.  We start with the limited labeled training set and build IC, which produces the embedding vectors of graph instances.  HC takes the embedding vectors as input and produces predictions.  We cautiously select a subset of predicted labels by HC with high confidence to enlarge the training set.  The enlarged training set is then fed into IC in the next iteration to update its parameters in the hope of generating more accurate embedding vectors and predictions.  HC further takes the new embedding vectors for model update and class prediction.  This is our proposed solution, called \underline{SE}mi-supervised gr\underline{A}ph c\underline{L}assification via \underline{C}autious \underline{I}teration (SEAL-CI), to the graph classification problem.

We also extend this iterative algorithm to the active learning framework, in which we iteratively select the most informative instances for manual annotation, and then update the classifiers with the newly labeled instances in a similar process as described above.  This method is called SEAL-AI in short.

\eat{
In all, in this work, we formulate a new problem named semi-supervised graph classification on a hierarchical graph. We propose a new framework, named \underline{SE}mi-superivsed gr\underline{A}ph c\underline{L}assification via \underline{C}autious/\underline{A}ctive Iteration (SEAL-C/AI), where "cautious" feeds to the scenario in which active learning is not available. In SEAL-C/AI, we first embed an arbitrary size of graph instance into a fixed-size embedding vector by the limited labeled instances at hand, then hierarchical graph is exploited. Based on the predictions of the two classifiers IC and HC, a candidate set is selected and committed to be "trustworthy". These "trustworthy" instances are utilized to update the graph embedding obtained in the previous iteration.

It is an iteration framework and the iteration finalizes if a pre-defined condition is certificated. To evaluate SEAL-C/AI, we implement it and apply it to both synthetic and real datasets. The real dataset is collected from the real online social network platform Tencent QQ, which has more than 800 million users and around 100 million online groups, from which we select nearly 40 thousand online groups and distinguish "game" groups from "non-game" groups.
}

Our contributions are summarized as follows.

\begin{itemize}
\item We study semi-supervised graph classification from a hierarchical graph perspective, which, to the best of our knowledge, has not been studied before.  Our proposed solutions SEAL-C/AI achieve superior classification performance to the state-of-the-other graph kernel and deep learning methods, even when given very few labeled training instances.

%We formulate a new problem named semi-supervised graph classification on a hierarchical graph, which is ubiquitous in graph-related data but seldomly explored by previous works. Our proposed framework SEAL-C/AI not only outperforms the potential alternatives by a significant margin in accuracy, but also provides meaningful interpretations of underlying knowledge.

\item We design a novel supervised, self-attentive graph embedding method called SAGE to embed graphs of arbitrary size into fixed-length vectors, which are used as a common form of input for classification.  The embedding approach not only simplifies the representation of a hierarchical graph greatly, but also provides meaningful interpretations of the underlying data in two forms: 1) embedding vectors of graph instances, and 2) node importance in a graph instance learned through a self-attentive mechanism that differentiates their contribution in classifying a graph instance.

\item We evaluate SEAL-C/AI on both synthetic graphs and Tencent QQ group data.  From the social networking platform Tencent QQ, we select 37,836 QQ groups with 18,422,331 unique anonymized users and classify them as ``game'' or ``non-game'' groups.  SEAL-C/AI achieve a Macro-F1 score of 70.8\% and 73.2\% respectively with only 2.6\% labeled instances.  They both outperform other competing methods by a large margin.
\end{itemize}

The remainder of this paper is organized as follows.  Section \ref{def} gives the problem definition and Section \ref{alt} describes the design of SEAL-C/AI.  We report the experimental results in Section \ref{sec.exp} and discuss related work in Section \ref{sec.related}.  Finally, Section \ref{sec.con} concludes the paper.

\begin{figure}
\begin{center}
\includegraphics [width=0.35\textwidth]{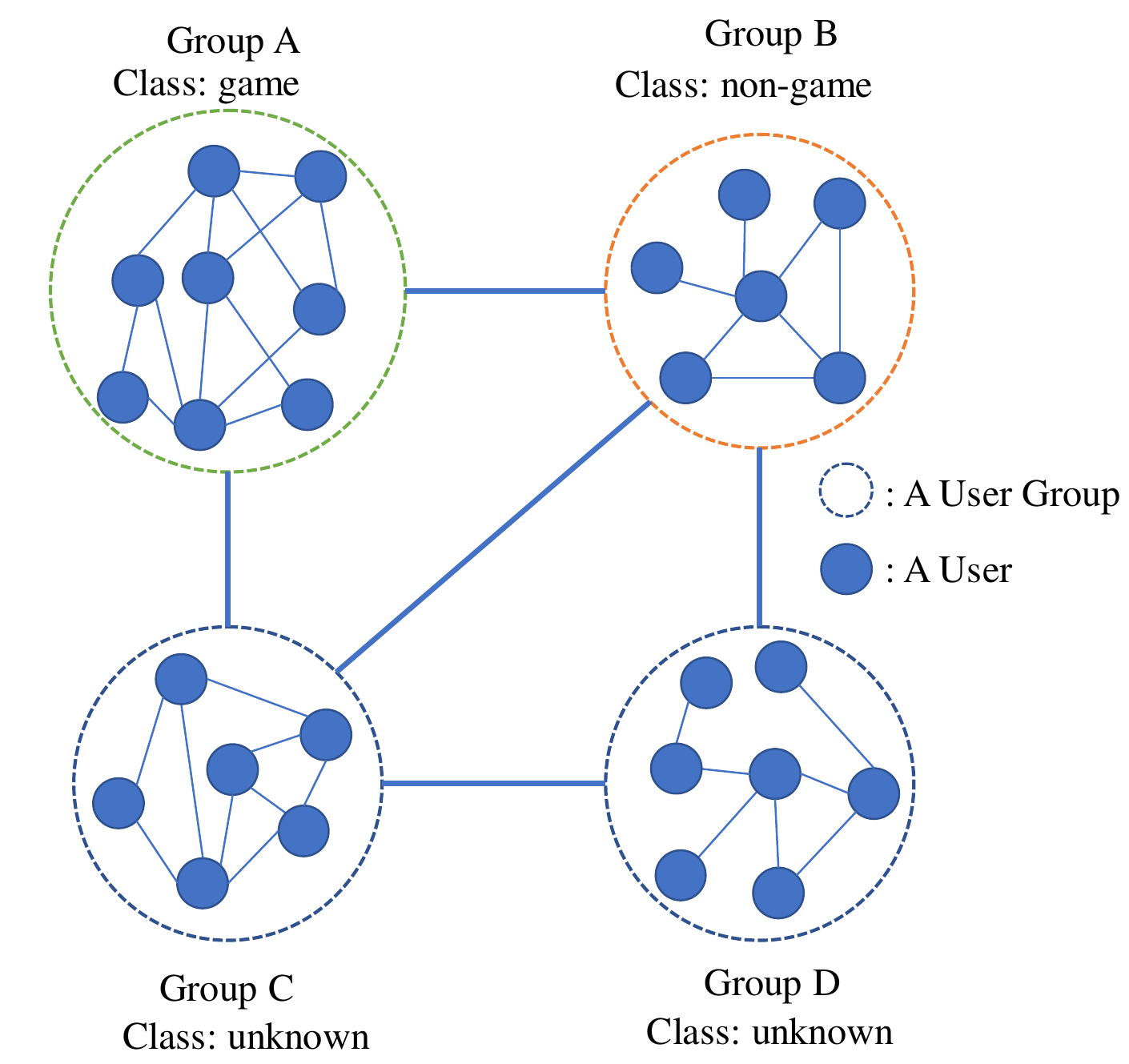}
\end{center}

\caption{A hierarchical graph with four graph instances $A, B, C, D$, each of which corresponds to a user group in a social network.}
\label{fig.example1}
\vspace{-0.3cm}
\end{figure}

\section{Problem Definition}\label{def}
We denote a set of objects as $O=\{o_1, o_2, \ldots, o_N\}$ which represent real-world entities.  We use $\phi$ attributes to describe properties of objects, e.g., age, gender, and other information of people.

We use a \emph{graph instance} to model the relationship between objects in $O$, which is denoted as $g=(V, A, X)$, $V\subseteq O$ is the node set and $|V|=n$, $A$ is an $n\times n$ adjacency matrix representing the connectivity in $g$, and $X \in \mathbb{R}^{n \times \phi}$ is a matrix recording the attribute values of all nodes in $g$.

A set of graph instances $G$ can be interconnected, and the connectivity between the graph instances is represented by an adjacency matrix $\Theta$.  The graph instances and their connections are modeled as a \emph{hierarchical graph}.

A graph instance $g\in G$ is a \emph{labeled graph} if it has a class label, represented by a vector $y\in \{0, 1\}^c$, where $c$ is the number of classes.  A graph instance is \emph{unlabeled} if its class label is unknown.  Then $G$ can be divided into two subsets: labeled graphs $G_l$ and unlabeled graphs $G_u$, where $G=G_l\cup G_u$, $|G_l|=L$ and $|G_u|=U$.  In this paper, we study the problem of \emph{\textbf{graph classification}}, which determines the class label of the unlabeled graph instances in $G_u$ from the available class labels in $G_l$ and the hierarchical graph topological structure.  As the amount of available class labels is usually very limited in real-world data, we take a semi-supervised learning approach to solving this problem.

Figure \ref{fig.example1} depicts a hierarchical graph in the context of a social network.  $A, B, C, D$ denote four user groups.  Group $A$ has the class label of ``game'', $B$ has the label of ``non-game'', while the class labels of $C$ and $D$ are unknown.  These four groups are connected via some kind of relationships, e.g., interactions or common members.  The internal structure of each user group shows the connections between individual users.  From this hierarchical graph, we want to determine the class labels of groups $C$ and $D$.

\section{Methodology}\label{alt}

\subsection{Problem Formulation}
In our problem setting, we have two kinds of information: graph instances and connections between the graph instances, which provide us with two perspectives to tackle the graph classification problem.  Accordingly, we build two classifiers: a classifier IC constructed for graph instances and a classifier HC constructed for the hierarchical graph, both of which make predictions for unlabeled graph instances in $G_u$.

For both classifiers, one goal is to minimize the supervised loss, which measures the distance between the predicted class probabilities and the true labels. Another goal is to minimize a disagreement loss, which measures the distance between the predicted class probabilities by IC and HC.  The purpose of this disagreement loss is to enforce a \emph{\textbf{consistency}} between the two classifiers.

\eat{
\begin{table}
  \caption{Notations}
  \label{Notations}
  \scalebox{1.0}{
  \begin{tabular}{ll}
    \toprule
    \textbf{Notation}&\textbf{Description} \\
    \midrule
	$\gamma$&output probabilities of GCN\\
	$\Gamma$&a set of output probabilities of GCN \\
	$\psi$&output probabilities of SAGE\\
	$\Psi$&a set of output probabilities of SAGE \\
	$e$&embedding vector for graph instance $g$ \\
	$\mathcal{W}$& set of all the parameters of SAGE and GCN\\
	$\Theta$& adjacency matrix that connects graph instances\\
  \bottomrule
\end{tabular}
}
\eat{\vspace{-0.3cm}}
\end{table}

}

Formally, we formulate the graph classification problem as an optimization problem:

\begin{equation}
 \min \zeta(G_l)+\xi(G_u),
\label{equ.total}
\end{equation}
where $\zeta(G_l)$ is the supervised loss for the labeled graph instances, and $\xi(G_u)$ is the disagreement loss for the unlabeled graph instances.

Specifically, $\zeta(G_l)$ includes two parts:
\begin{equation}
\zeta(G_l) = \sum_{g_i\in G_l}(\mathcal{L}(y_i, \psi_i) + \mathcal{L}(y_i, \gamma_i)),
\label{equ.super}
\end{equation}
where $\psi_i$ is a vector of predicted class probabilities by IC, and $\gamma_i$ is a vector of predicted class probabilities by HC.  $\mathcal{L}(\cdot, \cdot)$ is the cross-entropy loss function.

The disagreement loss $\xi(\cdot)$ is defined as:
\begin{equation}
\xi(G_u) = \sum_{g_i\in G_u}D_{KL}(\gamma_i || \psi_i),
\label{equ.unsuper}
\end{equation}
where $D_{KL}(\cdot||\cdot)$ is the Kullback-Leibler divergence, $D_{KL}(P||Q) = \sum_jP_j\log \big(\frac{P_j}{Q_j}\big)$.  In the following subsections, we describe our design of classifiers IC and HC, and our approach to minimizing the supervised loss and the disagreement loss.

\subsection{Design of Classifiers}\label{demCOCC}
Classifier IC takes a graph instance as input.  As different graph instances have different numbers of nodes, IC is expected to handle graph instances of arbitrary size. Classifier HC takes the hierarchical graph as input, in which individual graph instances are the ``nodes''.  This is a much too complicated input for a classifier.  To deal with the above challenges, we propose to embed a graph instance $g_i\in G$ into a fixed-length vector $e_i$ via IC first.  Then HC can take as input the embedding vectors of graph instances and the adjacency matrix $\Theta$. In particular, IC takes as input the adjacency matrix $A_i$ and attribute matrix $X_i$ of an arbitrary-sized graph instance $g_i$, and outputs an embedding vector $e_i$ and a vector of predicted class probabilities $\psi_i$, i.e., $(e_i, \psi_i) = \text{IC}(A_i, X_i)$.  HC takes the embedding vectors $E =\{e_i\}_{i=1}^{L+U}$ and $\Theta$, and outputs the predicted class probabilities $\Gamma = \{\gamma_i\}_{i=1}^{L+U}$, i.e., $\Gamma = \text{HC}(E, \Theta)$.  In the following, we illustrate the design of IC which performs discriminative graph embedding, and then the design of HC which performs graph-based classification.

\subsubsection{Discriminative graph embedding}\label{dem}

Our graph embedding task is to produce a fixed-length discriminative embedding vector of a graph instance.  In the literature, graph representation techniques have recently shifted from hand-crafted kernel methods~\cite{Yanardag:2015:DGK:2783258.2783417} to neural network based end-to-end methods, which achieve better performance in graph-structured learning tasks.  In this vein, we adopt neural network methods for the graph embedding task, for which, however, we identify three challenges:

\begin{itemize}
\item  \emph{Size invariance}: How to design the neural network structure to flexibly take an arbitrary-sized graph instance and produce a fixed-length embedding vector?
\item  \emph{Permutation invariance}: How to derive the representation regardless of the permutation of nodes?
\item  \emph{Node importance}: How to encode the importance of different nodes into a unified embedding vector?
\end{itemize}

In particular, the third challenge is \emph{node importance}, i.e., different nodes in a graph instance have different degrees of importance.  For example, in a ``game'' group the ``core'' members should be more important than the ``border'' members in contributing to the derived embedding vector.  We need to design a mechanism to learn the node importance and then encode it in the embedding vector properly.

\begin{figure*}
\begin{center}
\includegraphics [width=0.9\textwidth]{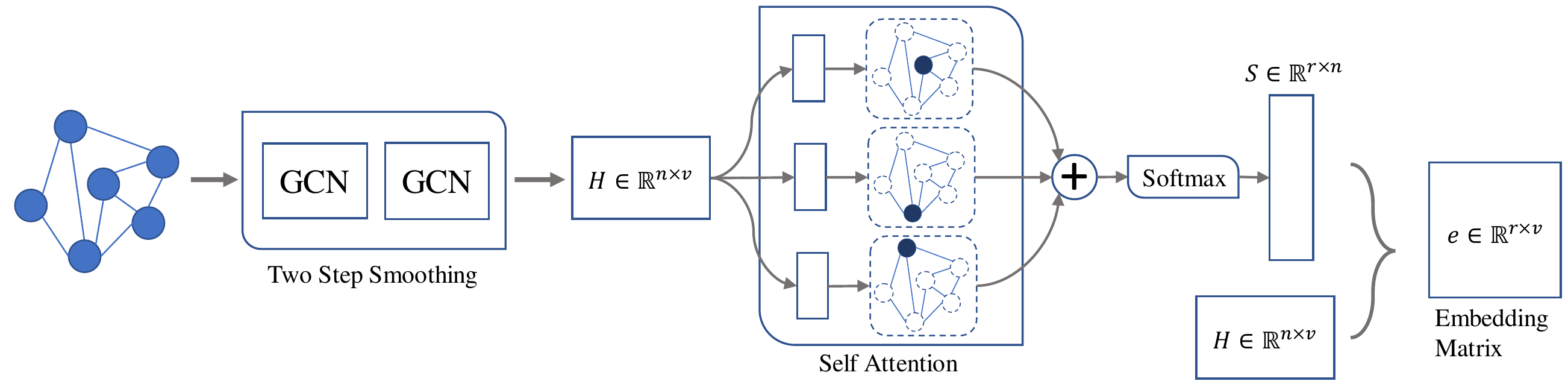}
\end{center}

\caption{The supervised self-attentive graph embedding method SAGE.}
\label{fig.SAGE}
\vspace{-0.3cm}
\end{figure*}

To this end, we propose a self-attentive graph embedding method, called SAGE, which can take a variable-sized graph instance, and combine each node to produce a fixed-length vector according to their importance within the graph.  In SAGE, we first utilize a multi-layer GCN~\cite{kipf2017semi} to smooth each node's features over the graph's topology.  Then we use a self-attentive mechanism to learn the node importance and then transform a variable number of smoothed nodes into a fixed-length embedding vector, as proposed in ~\cite{DBLP:journals/corr/LinFSYXZB17}.  Finally, the embedding vector is cascaded with a fully connected layer and a softmax function, in which the label information can be leveraged to discriminatively transform the embedding vector $e$ into $\psi$.  Figure~\ref{fig.SAGE} depicts the overall framework of SAGE.

Formally, we are given the adjacency matrix $A \in \mathbb{R}^{n \times n}$ and the attribute matrix $X \in \mathbb{R}^{n \times \phi}$ of a graph instance $g$ as input.  In the preprocessing step, the adjacency matrix $A$ is normalized:
\begin{equation}
  \hat{A} = \tilde{D}^{-\frac{1}{2}}(A + I_n)\tilde{D}^{-\frac{1}{2}},
\label{equ.gene}
\end{equation}
where $I_n$ is the identity matrix and $\tilde{D}_{ii} = \sum_m\ (A + I_n)_{im}$.  Then we apply a two-layer GCN network:
\begin{equation}
  H = \hat{A}\ \textsf{ReLU}(\hat{A}XW^0)W^1.
\label{equ.H}
\end{equation}
Here $W^0 \in \mathbb{R}^{\phi \times h}$ and $W^1 \in \mathbb{R}^{h \times v}$ are two weight matrices.  GCN can be considered as a Laplacian smoothing operator for node features over graph structures, as pointed out in~\cite{DBLP:journals/corr/abs-1801-07606}.  Then we get a set of representation $H \in \mathbb{R}^{n \times v}$ for nodes in $g$.  Note that the representation $H$ does not provide node importance, and it is size variant, i.e., its size is still determined by the number of nodes $n$.  So next we utilize the self-attentive mechanism to learn node importance and encode it into a unified graph representation, which is size invariant:
\begin{equation}
  S = \textsf{softmax} \big(W_{s2}\textsf{tanh}(W_{s1}H^T)\big),
\label{equ.gene}
\end{equation}
where $W_{s1} \in \mathbb{R}^{d \times v}$ and $W_{s2} \in \mathbb{R}^{r \times d}$ are two weight matrices.  The function of $W_{s1}$ is to linearly transform the node representation from a $v$-dimensional space to a $d$-dimensional space, then nonlinearity is introduced by tying with the function \textsf{tanh}. $W_{s2}$ is used as $r$ views of inferring the importance of each node within the graph. It acts like inviting $r$ experts to give their opinions about the importance of each node independently. Then \textsf{softmax} is applied to derive a standardized importance of each node within the graph, which means in each view the summation of all the node importance is 1.

After that, we compute the final graph representation $e \in \mathbb{R}^{r \times v}$ by multiplying $ S\in \mathbb{R}^{r \times n}$ with $H \in \mathbb{R}^{n \times v}$:
\begin{equation}
e = SH.
\end{equation}
$e$ is size invariant since it does not depend on the number of nodes $n$ any more.  It is also permutation invariant since the importance of each node is learned regardless of the node sequence, and only determined by the task labels.

One potential risk in SAGE is that $r$ views of node importance may be similar. To diversify their views of node importance, a penalization term is imposed:
\begin{equation}
  %P = \big|\big|\SS^T - I_r\big|\big|_F^2.
  P = \big|\big|\ SS^T - I_r\ \big|\big|_F^2.
\end{equation}
Here $\big|\big|\cdot\big|\big|_F$ represents the Frobenius norm of a matrix.  We train the classifier in a supervised way with the task at hand, in the hope of minimizing both the penalization and the cross-entropy loss.

To summarize, we use SAGE to construct the instance-level classifier IC.  It produces not only the estimated class probability vector $\psi$, but also a graph embedding $e$, which is the input for classifier HC described in the next subsection.

\subsubsection{Graph-based classification}\label{ssc}
Given the graph embedding $E = \{e_i\}_{i=1}^{L+U}$ and the adjacency matrix $\Theta \in \mathbb{R}^{(L+U) \times (L+U)}$, our next task is to infer the parameters of classifier HC and derive the predicted probabilities $\Gamma = \{\gamma_i\}_{i=1}^{L+U}$. \eat{It is done by solving problem \eqref{equ.super} while holding $E$ constant.} This problem falls into the setting of traditional graph-based learning where $E$ can be treated as the set of node features.  Recently neural network based approaches such as~\cite{kipf2017semi,yang2016revisiting} have demonstrated their superiority to traditional methods such as ICA~\cite{sen2008collective}.  In this context we make use of GCN~\cite{kipf2017semi} again for the consideration of efficiency and effectiveness. In the following, we consider a two-layer GCN and apply preprocessing by $\hat{\Theta} = \tilde{D_{\Theta}}^{-\frac{1}{2}}(\Theta + I_{L+U})\tilde{D_{\Theta}}^{-\frac{1}{2}}$. Then the model becomes:
\begin{equation}
  \Gamma = HC(E,\Theta) = \textsf{softmax} \big(\hat{\Theta}\ \textsf{ReLU}(\hat{\Theta}EW_{\Theta}^0)W_{\Theta}^1\big),
\label{Eq:GCN}
\end{equation}
where $W_{\Theta}^0 \in \mathbb{R}^{(rv) \times M}$ is an input-to-hidden weight matrix with $M$ feature maps and $W_{\Theta}^1 \in \mathbb{R}^{M \times c}$ is a hidden-to-output weight matrix. The $\textsf{softmax}$ function is applied row-wise and we get $\Gamma$.  With $\Gamma$ and $\Psi$ we can compute the supervised loss in problem \eqref{equ.super} and the disagreement loss in problem \eqref{equ.unsuper}.

\subsection{The Proposed SEAL-CI Model}\label{sec.method}
In this subsection, we present our method to minimize the objective function \eqref{equ.total}.  In real-world scenarios, the number of labeled graph instances $L$ can be quite small compared to the number of unlabeled instances $U$.  In this context, neural network based classifiers such as IC may suffer from the problem of overfitting.  To mitigate this, we have both the disagreement loss \eqref{equ.unsuper} and the supervised loss \eqref{equ.super} included in the objective function \eqref{equ.total}.  The disagreement loss can be regarded as a regularization to prevent overfitting.

Problem \eqref{equ.total} is a mixed combinatorial and continuous optimization problem.  The supervised loss \eqref{equ.super} includes two parts, $\mathcal{L}(y_i, \psi_i)$ and $\mathcal{L}(y_i, \gamma_i)$, i.e., the supervised loss of IC and HC.  $\mathcal{L}(y_i, \gamma_i)$ depends on classifier IC to provide accurate graph embedding.  All these issues make the problem highly non-convex.  As such, we use the idea of iterative algorithm to alternate minimizing the supervised loss of IC and HC, and minimizing the disagreement loss by trusting a subset of predictions by HC in the next iteration of graph embedding by IC.

To be more specific, we combine the graph embedding algorithm in Section \ref{dem} and graph-based classification algorithm in Section \ref{ssc} into one iterative algorithm.  We build IC to produce graph embedding $E^t$ for all graph instances in iteration $t$, and then feed $E^t$ into HC to get the predicted probabilities $\Gamma^t$.  We then make use of $\Gamma^t$ to update the parameters of IC and generate $E^{t+1}$, which is then used as the input of HC in iteration $t+1$.  Figure~\ref{fig.em} depicts the overall framework of this iterative process.  Although this method may not reach the global optimum, similar setting~\cite{mcdowell2007cautious,sen2008collective} has been proven to be effective.
\begin{figure*}
\begin{center}
\includegraphics [width=1\textwidth]{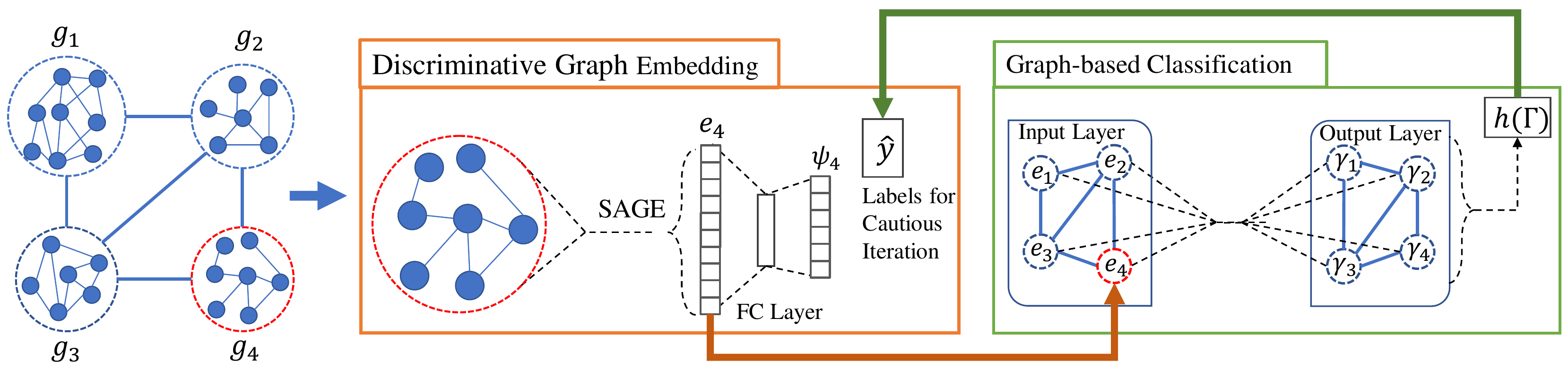}
\label{fig.em}
\end{center}

\caption{Schematic diagram of the learning framework SEAL-CI. There are two subroutines: discriminative graph embedding (in the orange box) and graph-based classification (in the green box).}
\label{fig.em}
\vspace{-0.3cm}
\end{figure*}

\begin{algorithm}[t]
  \caption{SEAL-CI}
  \label{algo.naive}
  \KwIn{$A$, $X$, $\Theta$.}
  \KwOut{$\Psi^t$, $\Gamma^t$.}
  Initial: $G_{tmp} =\emptyset$, $G_l^0 = G_l$, $t = 0$;

  \While{$t\lambda \leq U$}{
    $\mathcal{W}^{t+1} \leftarrow$ $\arg\min\zeta(G_l^{t}| \mathcal{W}^{t})$\;

    $ \Psi^{t+1}, E^{t+1} \leftarrow$ \textsf{IC}($A, X | \mathcal{W}^{t+1}$)\;
    $ \Gamma^{t+1} \leftarrow$ \textsf{HC}($E^{t+1}, \Theta | \mathcal{W}^{t+1}$)\;
    %\While{$|G_{tmp}| < \lambda$ and $G_u^{t} \neq \emptyset$}{
    %    $g_i = \arg\max_{g_i \in G_u^t, \gamma_{g_i} \in \Gamma^{t+1}}\gamma_{g_i}$\;
    %    $G_{tmp} \leftarrow g_i$\;
    %    $G_u^{t} \leftarrow G_u^{t} \setminus g_i$\;
    %}
	$G_{tmp} \leftarrow h(t\lambda,\Gamma_{G_u}^{t+1} )$\;
	%$G_u^{t+1} \leftarrow G_u^{t} \setminus G_{tmp}$\;
    $G_l^{t+1} \leftarrow G_l \cup G_{tmp}$\;
    %$G_u^{t} \leftarrow G_u^{t+1}$\;
	$G_{tmp} =\emptyset$\;
	%$t \leftarrow t + 1$\;
  }
  Return $\Psi^t$, $\Gamma^t$\;
\end{algorithm}

\subsubsection{How to utilize $\Gamma^t$?}
To update the graph embedding vectors, a naive approach is feeding the whole set of $\Gamma^t$ for the parameter update in IC, which is the idea of the original ICA~\cite{sen2008collective}.  However, not all $\Gamma^t$ are correct in their predictions.  The false predictions may lead the update of embedding neural network to the wrong direction. \eat{Within the set of unlabeled graphs, different $\gamma$ could contribute differently to the update of embedding neural network.} To this end, we make use of the idea of \cite{mcdowell2007cautious}, a variant of the original ICA, and cautiously exploit a subset of $\Gamma^t$ to update the parameters of IC in each iteration.  Specifically, in iteration $t$, we choose the $t\lambda$ most confident predicted labels while ignoring the less confident predicted labels.  This operation continues until all the unlabeled samples have been utilized.  To further improve the efficiency, the parameters of IC are not re-trained but fine-tuned based on the parameters obtained in the previous iteration.  This algorithm is called \underline{SE}mi-supervised gr\underline{A}ph c\underline{L}assification via \underline{C}autious \underline{I}teration (SEAL-CI) and is presented in Algorithm ~\ref{algo.naive}.  Note here $\mathcal{W}$ is the set of all the parameters of IC and HC. \eat{In line $5$, we are sorting $\Gamma$ by its maximum probability.} In line $6$, the training set for IC has been enlarged by $t\lambda$ instances and it is done by ``committing'' these instances' labels from their maximum probability.  In other words, the newly enrolled training instances are found by:
\begin{equation}
  h(\lambda,\Gamma) = \textsf{top}(\max_{\gamma \in \Gamma} \gamma,\lambda).
\end{equation}
Here function $\textsf{top}(\cdot,\lambda)$ is used to select the top $\lambda$ instances and function $\max\gamma$ is used to select the maximum value in the probability vector $\gamma$.  \eat{By ``committing'' these instances, we turn a probability vector into its one-hot form.  For example, it transforms a vector of $(0.1,0.2,0.3,0.4)$ into $(0,0,0,1)$.}

\subsection{The Proposed SEAL-AI Model}\label{sec.ai}
Our proposed model is easy to extend to the active learning scenario.  In case further annotation is available, we can perform active learning and ask for annotations with a budget of $B$.  Denote the set of graph instances being annotated as $G_B$, then the objective function in the active learning setting is re-written as:
\begin{equation}
\begin{split}
  \min f(G|B, \mathcal{W})\\
  \text{s.t.}\ \ \ |G_B| \leq B,
\end{split}
\label{equ.ctotal}
\end{equation}
where $f(G|B, \mathcal{W}) = \zeta(G_l \cup G_B | \mathcal{W})+\xi(G_u \setminus G_B | \mathcal{W})$.  This is still a mixed combinatorial and continuous optimization problem.  It is very hard to infer the model parameters and the active learning set $G_B$ simultaneously.  By definition, the active learning set $G_B$ is intractable unless the model parameters are completely inferred.  To solve this chicken-and-egg problem, we decompose the objective function into two sub-steps: parameter optimization and candidate generation. Then we optimize $f(G|B, \mathcal{W})$ iteratively.  This algorithm is called \underline{SE}mi-supervised gr\underline{A}ph c\underline{L}assification via \underline{A}ctive \underline{I}teration (SEAL-AI) and is shown in Algorithm~\ref{algo.set}.

\begin{algorithm}[t]
  \caption{SEAL-AI}
  \label{algo.set}
  \KwIn{$A$, $X$, $\Theta$.}
  \KwOut{$\Psi^t$, $\Gamma^t$.}
  %Initial: $E^0 \leftarrow$ graph2vec \;
  Initial: $G_{tmp} =\emptyset$,$G_B^0 = \emptyset$, $G_l^0 = G_l$, $G_u^0 = G_u$, $t = 0$;

  \While{$|G_B^t|$ $\leq$ $B$ }{
    %// Parameter optimization

    $\mathcal{W}^{t+1} \leftarrow$ $\arg\min\zeta(G_l^{t}| \mathcal{W}^{t})$\;
	$ \Psi^{t+1}, E^{t+1} \leftarrow$ \textsf{IC}($A, X | \mathcal{W}^{t+1}$)\;
    $ \Gamma^{t+1} \leftarrow$ \textsf{HC}($E^{t+1}, \Theta| \mathcal{W}^{t+1}$)\;
    %// Candidate generation

    $G_{tmp}$ $\leftarrow$ $\arg\min_{|G_{tmp}| = k}\xi(G_u^{t} \setminus G_{tmp} | \mathcal{W}^{t+1})$\;
    $G_B^{t+1} \leftarrow G_B^{t} \cup G_{tmp}$\;
    $G_l^{t+1} \leftarrow G_l^{t} \cup G_{tmp}$\;
    $G_u^{t+1} \leftarrow G_u^{t} \setminus G_{tmp}$\;
	$G_{tmp} =\emptyset$\;
  }
  Return $\Psi^t$, $\Gamma^t$\;
\end{algorithm}

At the beginning of this iterative process, we optimize the supervised loss $\zeta(G_l| \mathcal{W})$ based on current labeled graphs in $G_l$ (line 3 in Algorithm~\ref{algo.set}). In active learning, the choice of candidate generator is a key component.  We exploit the idea of ALFNET~\cite{bilgic2010active} and choose the candidate graph instances $G_{tmp}$ by maximizing the decrease of the current disagreement loss based on the new parameter obtained in the first step (line 6 in Algorithm~\ref{algo.set}).  At last we label $G_{tmp}$ and update $G_B$, $G_l$ and $G_u$ respectively (line 7-9 in Algorithm~\ref{algo.set}).  \eat{Since $\zeta(G_l^{t}| \mathcal{W}^{t}) \geq \zeta(G_l^{t}| \mathcal{W}^{t+1})$ and $\xi(G_u^{t}|\mathcal{W}^{t+1}) \geq \xi(G_u^{t+1}|\mathcal{W}^{t+1})$,  Algorithm~\ref{algo.set} can converge in finite steps.}

It is worth noting that from the hard example mining perspective, the disagreement score is an excellent criterion for the active learning setting. Specifically, we choose the candidates by first calculating the distribution divergence of $(\gamma_i,\psi_i)$ from $\Gamma_u = \{\gamma_i\}_{i=1}^U$ and $\Psi_u = \{\psi_i\}_{i=1}^U$:
\begin{equation}
  z(\psi_i,\gamma_i) = D_{KL}(\gamma_i || \psi_i).
\end{equation}
Then we choose $k$ instances with the largest KL divergence.  Intuitively, the KL divergence between $\psi_i$ and $\gamma_i$ can be viewed as the conflict of two supervised models.  A large KL divergence indicates that one of the models gives wrong predictions. To this end, the instances with a large KL divergence are more informative to help the algorithm converge more quickly.

\subsection{Complexity Analysis}
We analyze the computational complexity of our proposed methods.  Here we only focus on Algorithm~\ref{algo.naive}, since Algorithm ~\ref{algo.set} is almost the same except the step of selecting candidate graph instances to the training set.  In Algorithm~\ref{algo.naive}, the intensive parts in each iteration contain the updates of IC and HC as well as the selection of candidate instances. We discuss each part in details below.

Regarding IC, the core is to compute the activation matrix $H$ in Eq.~\eqref{equ.H} where the matrix-vector multiplications are up to $O(E_1\phi)$ flops for one input graph instance; here $E_1$ denotes the number of edges in the graph instance and $\phi$ is the input feature dimension.  Thus, it leads to the complexity of $O(E_1(L+U)\phi)$ by going through all $L+U$ graph instances.

Next, the computation by HC in Eq.~\eqref{Eq:GCN} requires $O(E_2rv)$ flops in total, where $E_2$ denotes the number of links between graph instances.  Then in candidate selection, performing comparisons between all unlabeled graph instances has a complexity of $O(L+U)$ given the outputs of two classifiers IC and HC.

Overall, the complexity of our method is $O(E_1(L+U)\phi+E_2rv)$ which scales linearly in terms of the number of edges in each graph instance (i.e., $E_1$), the number of links between graph instances (i.e., $E_2$) and the number of graph instances (i.e., $(L+U)$).  Thus, our method is computationally comparable to the GCN-based method~\cite{kipf2017semi}, and more efficient than PSCN~\cite{Niepert2016LearningCN} that is quasi-linear with respect to the numbers of nodes and edges.

\section{EXPERIMENTS}\label{sec.exp}
We first validate the effectiveness of our graph embedding algorithm SAGE on two data sets: PROTEINS and D\&D.  Then we evaluate our SEAL-C/AI methods on both synthetic and Tencent QQ group data sets.

\subsection{Performance of SAGE}
We use two benchmark data sets, PROTEINS and D\&D, to evaluate the classification accuracy of SAGE, and compare it with the state-of-the-art graph kernels and deep learning approaches.  PROTEINS \cite{borgwardt2005protein} is a graph data set where nodes are secondary structure elements and edges represent that two nodes are neighbors in the amino-acid sequence or in 3D space.  D\&D \cite{dobson2003distinguishing} is a set of structures of enzymes and non-enzymes proteins, where nodes are amino acids, and edges represent spatial closeness between nodes.  Table \ref{tab:twodatasets} lists the statistics of these two data sets.

\begin{table}
  \caption{Statistics of PROTEINS and D\&D}
  \label{tab:twodatasets}
  \begin{tabular}{ccc}
    \toprule
&\textbf{PROTEINS}& \textbf{D\&D}\\
    \midrule
	Max number of nodes &620&5748\\
	Avg number of nodes &39.06&284.32\\
	Number of graphs &1113&1178\\
  \bottomrule
\end{tabular}
\end{table}

\subsubsection{Baselines and Metrics}\label{bench.base}

The baselines include four graph kernels and two deep learning approaches:

\begin{itemize}
\item the shortest-path kernel (SP) \cite{borgwardt2005shortest},

\item the random walk kernel (RW) \cite{gartner2003graph},

\item the graphlet count kernel (GK) \cite{shervashidze2009efficient},

\item the Weisfeiler-Lehman subtree kernel (WL) \cite{shervashidze2011weisfeiler},

\item PATCHY-SAN (PSCN) \cite{Niepert2016LearningCN}, and

\item graph2vec \cite{DBLP:journals/corr/NarayananCVCLJ17}.

\end{itemize}

We follow the experimental setting as described in \cite{Niepert2016LearningCN}, and perform 10-fold cross validation.  In each partition, the experiments are repeated for 10 times.  The average accuracy and the standard deviation are reported.  We list results of the graph kernels and the best reported results of PSCN according to \cite{Niepert2016LearningCN}.

For SAGE, we use the same network architecture on both data sets.  The first GCN layer has 128 output channels, and the second GCN has 8 output channels.  We set $d=64$, $r=16$, and the penalization term coefficient to be $0.15$.  The dense layer has 256 rectified linear units with a dropout rate of 0.5. We use minibatch based Adam \cite{DBLP:journals/corr/KingmaB14} to minimize the cross-entropy loss and use He-normal \cite{he2015delving} as the initializer for GCN.  For both data sets, the only hyperparameter we optimized is the number of epochs.

\subsubsection{Results}

Table \ref{tab:sasc} lists the experimental results.  As we can see, SAGE outperforms all the graph kernel methods and the two deep learning methods by 1.27\% -- 5.59\% in accuracy.  This shows that our graph embedding method SAGE is superior. %In addition, since SAGE does not need any preprocessing such as kernel matrix calculation in kernel methods or node ordering in PSCN, it is much faster than these competitors \textbf{(to be done)}.

\begin{table}
  \caption{Accuracy of different classifiers}
  \label{tab:sasc}
  \begin{tabular}{ccc}
    \toprule
    \textbf{Approach}&\textbf{PROTEINS}& \textbf{D\&D}\\
    \midrule
	SP&75.07\textpm 0.54\%&-\\
	RW&74.22\textpm 0.42\%&-\\
	GK&71.67\textpm 0.55\%&78.45\textpm 0.26\%\\
	WL&72.92\textpm 0.56\%&77.95\textpm 0.70\%\\
	PSCN&75.89\textpm 2.76\%&77.12\textpm 2.41\%\\
	graph2vec&73.30\textpm 2.05\%&-\\
	SAGE&\textbf{77.26}\textpm 2.28\%&\textbf{80.88}\textpm 2.33\%\\
  \bottomrule
\end{tabular}
\vspace{-0.3cm}
\end{table}

\subsection{SEAL-C/AI on Synthetic Data}

We evaluate the performance of SEAL-C/AI on synthetic data.  We first give a description of the synthetic generator, then visualize the learned embeddings and analyze the self-attentive mechanism on the generated data.  Finally we compare our methods with baselines in terms of classification accuracy.

\subsubsection{Synthetic Data Generation}

The benchmark data set Cora \cite{mccallum2000automating} contains 2708 papers which are connected by the ``citation'' relationship.  We borrow the topological structure of Cora to provide the skeleton (i.e., edges) of our synthetic hierarchical graph.  Then we generate a set of graph instances, which serve as the nodes of this hierarchical graph.  Since there are 7 classes in Cora, we adopt 7 different graph generation algorithms, that is, Watts-Strogatz \cite{watts1998collective}, Tree graph, Erd{\H o}s-R{\'e}nyi \cite{erdos1960evolution}, Barbell \cite{herbster2007prediction}, Bipartite graph, Barab$\acute{a}$si-Albert graph \cite{bollobas2003mathematical} and Path graph, to generate 7 different types of graph instances, and connect them in the hierarchical graph.

Specifically, to generate a graph instance $g$, we randomly sample a number from $[100, 200]$ as its size $n$.  Then we generate its structure and assign the class label according to the graph generation algorithm.  In this step, the parameter $p$ in Watts-Strogatz, Erd{\H o}s-R{\'e}nyi, Bipartite graph and Barab$\acute{a}$si-Albert graph is randomly sampled from $[0.1, 0.5]$, the branching factor for Tree graph is randomly sampled from $[1, 3]$.  At last, to make this problem more challenging, we randomly remove $1\%$ to $20\%$ edges in the generated graph $g$.  The statistics of the generated graph instances are listed in Table \ref{tab:sgg}.

\begin{table}
  \caption{Statistics of generated graph instances}
  \label{tab:sgg}
  \scalebox{1.0}{
  \begin{tabular}{ccccc}
    \toprule
    \textbf{Type}&\textbf{Number}&\textbf{Nodes}&\textbf{Edges}&\textbf{Density} \\
    \midrule
	Watts-Strogatz&351&173&347&2.3\%\\
	Tree&217&127&120&1.5\%\\
	Erd{\H o}s-R{\'e}nyi&418&174&3045&20\%\\
	Barbell&818&169&2379&16.3\%\\
	Bipartite&426&144&1102&10.6\%\\
	Barab$\acute{a}$si-Albert&298&173&509&3.4\%\\
	Path&180&175&170&1.1\%\\
  \bottomrule
\end{tabular}
}
\raggedright{The node and edge numbers and density are the average for each type of graph.}
\vspace{-0.3cm}
\end{table}

\subsubsection{Visualization}
\begin{figure}
\begin{center}
\includegraphics [width=0.4\textwidth,scale=1]{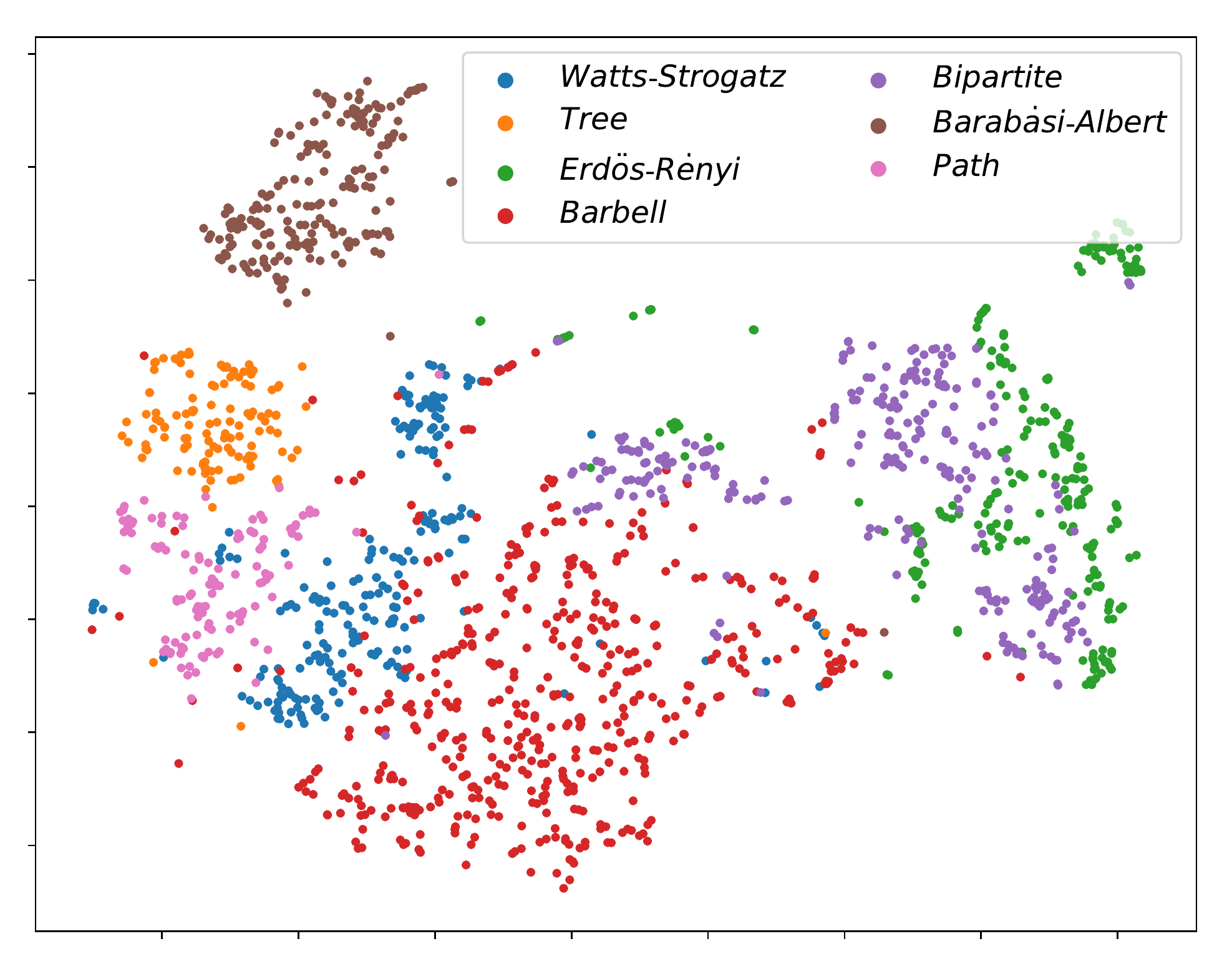}
\end{center}

\caption{Two-dimensional visualization of graph embeddings generated from the synthesized graph instances using SAGE. The nodes are colored according to their graph types.}
\label{fig.diary}
\vspace{-0.3cm}
\end{figure}

To have a better understanding of the synthesized graph instances, we split all 2708 graph instances into two parts.  1708 instances are used for training and the remaining 1000 instances are used for testing.  We apply SAGE on the training set and derive the embeddings of the 1000 testing instances.  We then project these learned embeddings into a two-dimensional space by t-SNE \cite{maaten2008visualizing}, as depicted in Figure \ref{fig.diary}. Each color in Figure \ref{fig.diary} represents a graph type.  As we can see from this two-dimensional space, the geometric distance between the graph instances can reflect their graph similarity properly.

\begin{figure*}
\centering
\includegraphics [width=0.28\textwidth]{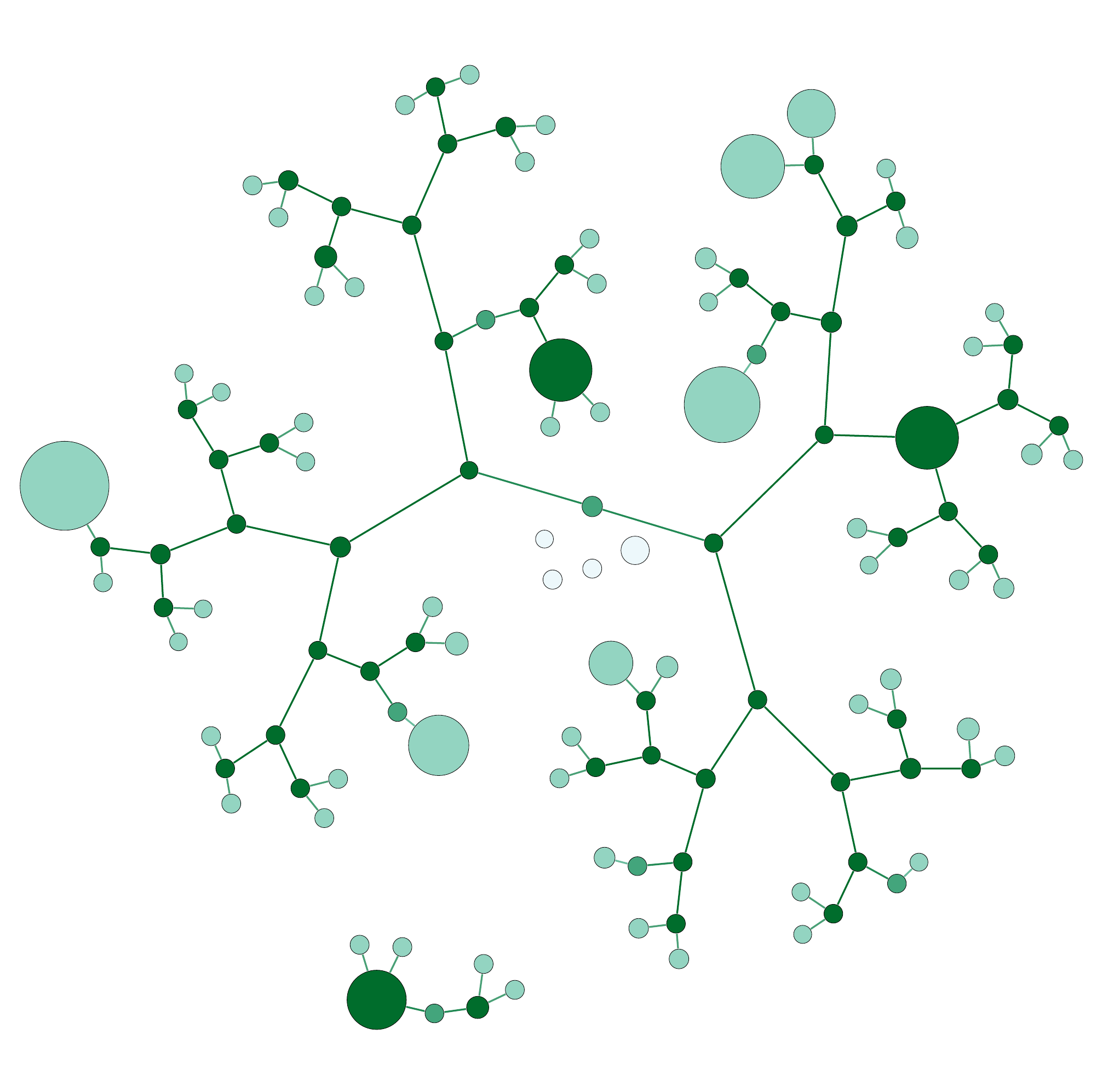}
\includegraphics [width=0.28\textwidth]{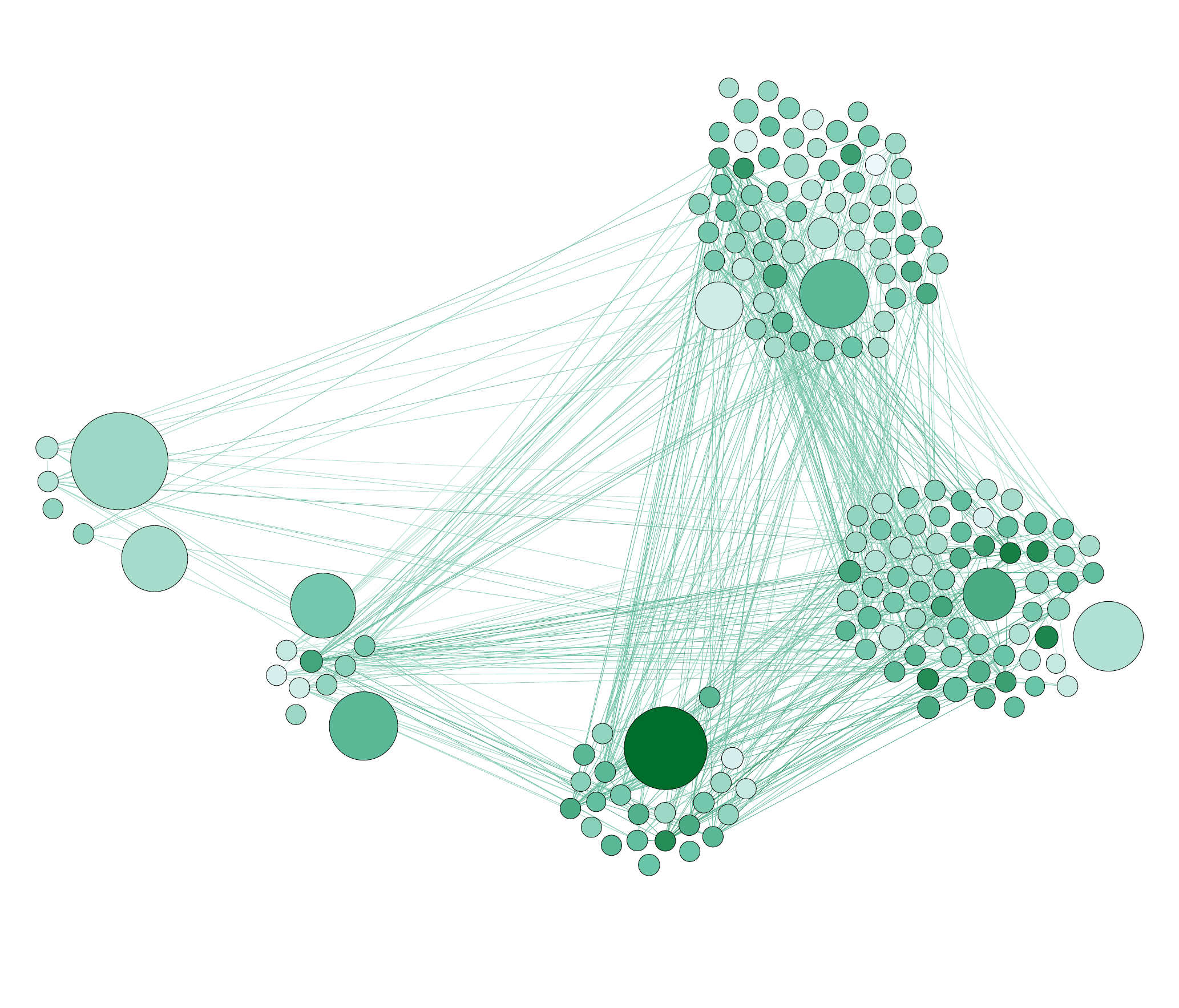}
\includegraphics [width=0.28\textwidth]{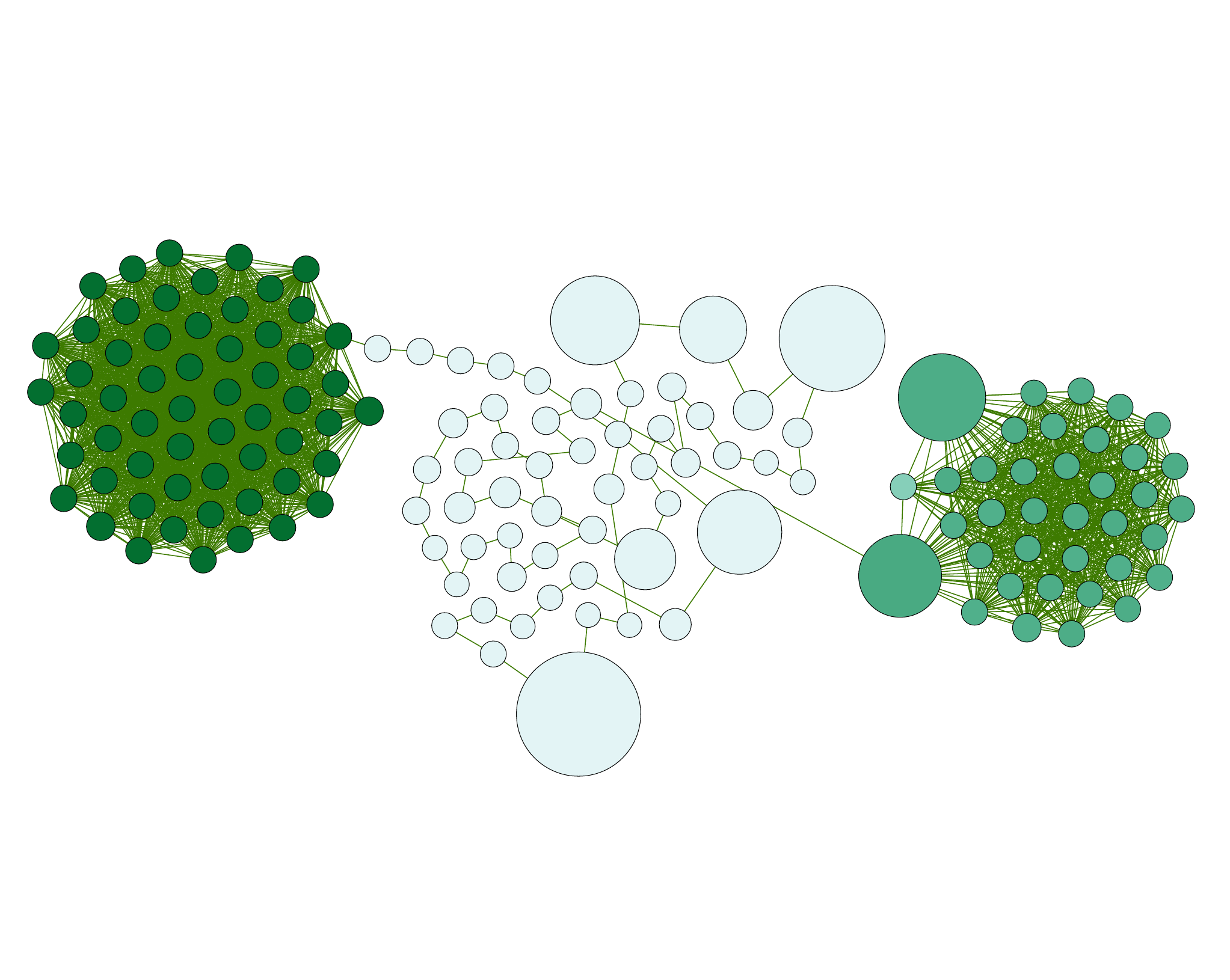}
\label{rhythm}
\caption{Attention of graph embeddings on 3 different types of graphs (left: Tree graph; middle: Erd{\H o}s-R{\'e}nyi graph; right: Barbell graph). A bigger node indicates a larger importance, and a darker color implies a larger node degree.}
\label{fig.rhythm}
\end{figure*}

We then examine the self-attentive mechanism of SAGE.  We calculate the average attention weight across $r$ views and normalize the resulting attention weights to sum up to 1. From the testing instances, we select three examples: a Tree graph, an Erd{\H o}s-R{\'e}nyi graph and a Barbell graph, for which SAGE has a high confidence ($>0.9$) in predicting their class label.  The three examples are depicted in Figure \ref{fig.rhythm}, where a bigger node implies a larger average attention weight, and a darker color implies a larger node degree.  On the left is a Tree graph, in which most of the important nodes learned by SAGE are leaf nodes.  This is reasonable since leaves are discriminative features to distinguish Tree graph from the other 6 types of graphs.  In the center is an Erd{\H o}s-R{\'e}nyi graph. We cluster these nodes into 5 groups by hierarchical clustering \cite{johnson1967hierarchical}, and see that SAGE tends to highlight those nodes with large degrees within each cluster.  On the right is a Barbell graph, in which SAGE pays attention to two kinds of nodes.  The first kind is those nodes that connect a dense graph and a path, and the second kind is the nodes that are on the path.

\subsubsection{Baselines and Metrics}\label{syn.base}
We use 6 approaches as our baselines:
\begin{itemize}
\item GK-SVM/GCN \cite{shervashidze2009efficient}, which calculates the graphlet count kernel (GK) matrix, then GK-SVM feeds the kernel matrix into SVM \cite{hearst1998support} whereas GK-GCN feeds the kernel vector of each graph instance to GCN.

\item WL-SVM/GCN \cite{shervashidze2011weisfeiler}, which is similar as above but using the Weisfeiler-Lehman subtree kernel (WL).

\item graph2vec-GCN \cite{DBLP:journals/corr/NarayananCVCLJ17}, which embeds the graph instances by graph2vec and then feeds the embeddings to GCN.

\item cautious-SAGE-Cheby, which is similar to SEAL-CI except that we replace GCN with Cheby-GCN \cite{defferrard2016convolutional}.

\item active-SAGE-Cheby, which is similar to SEAL-AI except that we replace GCN with Cheby-GCN \cite{defferrard2016convolutional}.

\item SAGE, which ignores the connections between graph instances and treats them independently.
\end{itemize}

\eat{For the first two approaches, we also use SVM \cite{hearst1998support} to classify these graph instances on the computed kernel matrix and report the better performance from SVM and GCN.

}
We use 300 graph instances as the training set for all methods except SEAL-AI and active-SAGE-Cheby, for which only 140 graphs are used as labeled graph instances at hand and then $B =160$ is set for active learning.  We use 1000 graph instances as the testing set.  We run each method 5 times and report its average accuracy.  The number of epochs for graph2vec is 1000 and the learning rate is 0.3.  To avoid overfitting of SAGE on this small data set, we use a relatively small number of neurons.  The first GCN layer has 32 output channels and the second GCN layer has 4 output channels.  We set $d=32$ and $r=10$.  The dense layer has 48 units with a dropout rate of 0.3. We set $M = 16$ in HC.

\subsubsection{Results}
Table \ref{tab:synr} shows the experimental results for semi-supervised graph classification.  Among all approaches, SEAL-C/AI achieve the best performance.  In the following, we analyze the performance of all methods categorized into 4 groups.

\noindent\underline{Group *1}: Both GK-SVM and WL-SVM outperform their GCN-based counterparts, indicating that SVM is more effective than GCN with the computed kernel matrix.  All the embedding-based methods perform better than these two kernel methods, which proves that embedding vectors are effective representations for graph instances and are suitable input for graph neural networks.

\noindent\underline{Group *2}: graph2vec-GCN achieves 85.2\% accuracy, which is comparable to that of SAGE, but lower than that of SEAL-C/AI.  One possible explanation is that graph2vec is an unsupervised embedding method, which fails to generate discriminative embeddings for classification.  Another possibility is that there is no iteration in this method, and the 300 training instances do not include very informative ones. These limitations of graph2vec are also motivations for us to design the supervised embedding method SAGE and the iterative framework in SEAL-CI.

\noindent\underline{Group *3}: cautious-SAGE-Cheby outperforms SAGE by only 0.8\%, which is not remarkable considering that it exploits many more training instances.  The accuracy of active-SAGE-Cheby is 3.3\% lower than that of SEAL-AI, which means that Cheby-GCN is inferior to GCN.

\noindent\underline{Group *4}: Both SEAL-CI and SEAL-AI outperform SAGE significantly, which proves the effectiveness of our hierarchical graph based perspective and the iterative algorithm for graph classification.  SEAL-AI outperforms SEAL-CI only slightly, by 1.2\%.  This shows, although SEAL-CI can make use of more training samples, it is still influenced by the misclassified cases of GCN.

\begin{table}
  \caption{Comparison of different methods on the synthetic data set for semi-supervised graph classification}
  \label{tab:synr}
  \scalebox{0.9}{
  \begin{tabular}{ccc}
    \toprule
     &\textbf{Algorithm}&\textbf{Accuracy} \\
    \midrule
		\multirow{2}{*}{*1}& \textbf{GK-SVM/GCN} & 77.8\%/73.4\%\\
		& \textbf{WL-SVM/GCN} & 83.4\%/75.5\%\\
		\hline
		\multirow{1}{*}{*2} & \textbf{graph2vec-GCN} &85.2\%\\
		\hline
		\multirow{2}{*}{*3}& \textbf{cautious-SAGE-Cheby} & 86.5\%\\
		& \textbf{active-SAGE-Cheby} & 89.1\%\\
		\hline
		\multirow{3}{*}{*4} & \textbf{SAGE} & 85.7\% \\
		& \textbf{SEAL-CI} & 91.2\%\\
		& \textbf{SEAL-AI} & \textbf{92.4}\%\\
	  \bottomrule
\end{tabular}
}
\eat{\vspace{-0.4cm}}
\end{table}

\subsubsection{Influence of the number of labeled training instances}
\begin{figure}
\begin{center}
\includegraphics [width=0.4\textwidth,scale=1]{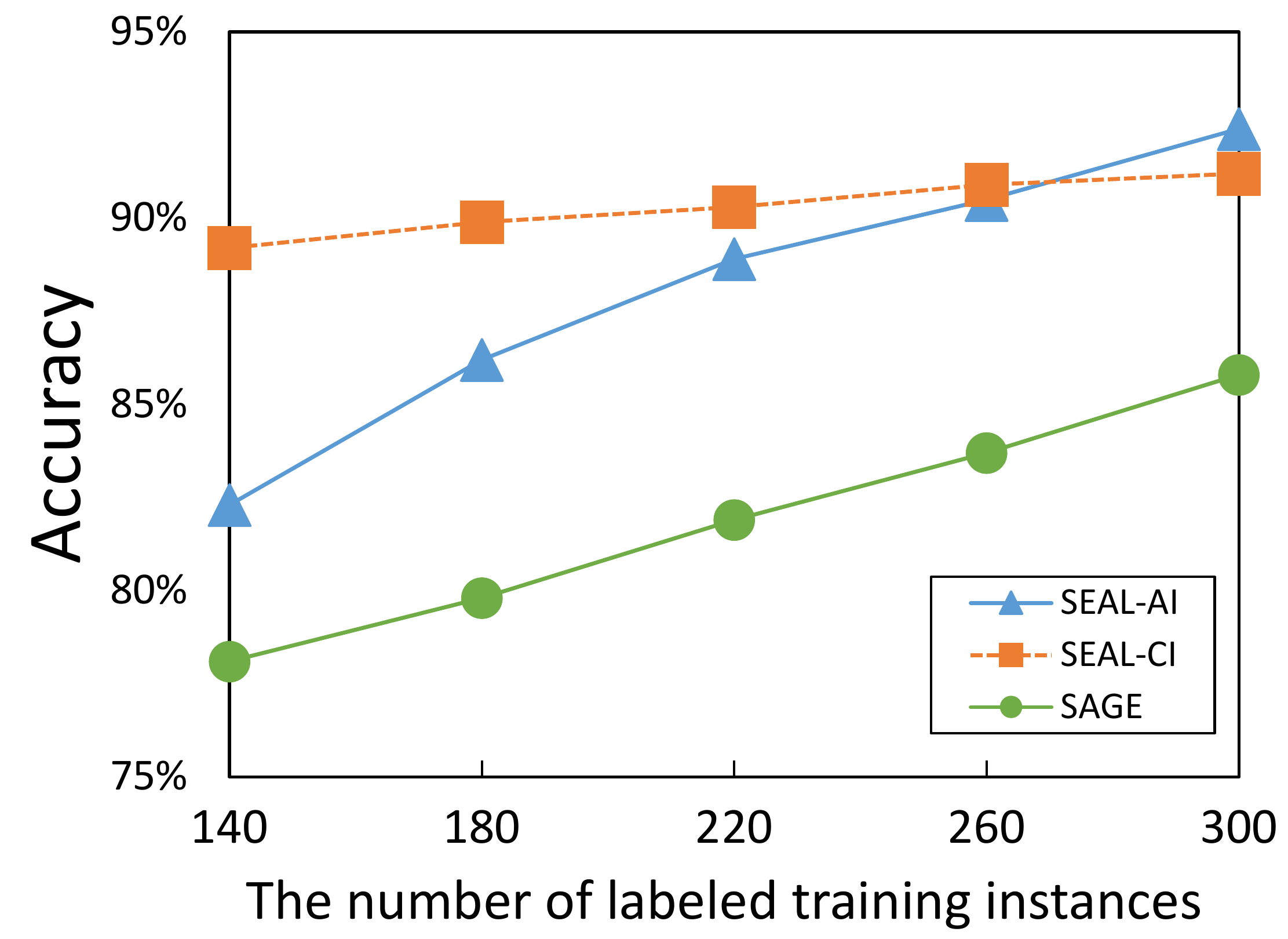}
\end{center}

\caption{Accuracy with different number of labeled training instances on synthetic data for semi-supervised graph classification.}
\label{fig.bud}
\vspace{-0.3cm}
\end{figure}

We examine how the number of labeled training instances affects the performance of our methods.  We train SAGE and SEAL-CI with a label size of $\{140, 180, 220, 260, 300\}$.  We train SEAL-AI with 140 labeled instances and then set the budget $B$ for active learning at $\{0, 40, 80, 120, 160\}$.  Thus the three methods have the same number of labeled training instances.  We set $\lambda=40$ in SEAL-CI and $k=10$ in SEAL-AI.  We run all methods 5 times, and plot their average accuracy in Figure \ref{fig.bud}.  As we can see from Figure \ref{fig.bud}, when the number of labeled training instances is 140, SEAL-CI performs best since it can utilize more training samples.  As the number of labeled training instances increases, the performance of SEAL-AI improves dramatically.  SEAL-AI catches up with SEAL-CI at 260 labeled training instances and outperforms SEAL-CI at 300 labeled training instances.  It validates that SEAL-AI can make use of the iterations to find informative and accurate training samples.  Meanwhile SEAL-CI trusts the prediction of GCN conditionally on its confidence, which may bring some noise to the learning process.  SEAL-C/AI outperform SAGE in all cases, which makes sense because SEAL-C/AI make good use of the hierarchical graph setting and consider the connections between the graph instances for classification.

\subsection{SEAL-C/AI on Tencent QQ Group}
In this section, we evaluate SEAL-C/AI on Tencent QQ group data.  We describe the characteristics of this data set and then present the experimental results.  Finally, we have some open discussions on how to construct a hierarchical graph from real-world data.

\subsubsection{Data Description}
Tencent QQ is a social networking platform in China with nearly 800 million monthly active users\footnote{\url{https://www.tencent.com/en-us/articles/17000391523362601.pdf}}.  There are around 100 million active online QQ groups. In this experiment, we select 37,836 QQ groups with 18,422,331 unique anonymized users.  For each user, we extract seven personal features:
\begin{itemize}
\item number of days ever since the registration day;
\item most frequently active area code in the past 90 days;
\item number of friends;
\item number of active days in the past 30 days;
\item number of logging in the past 30 days;
\item number of messages sent in the past 30 days;
\item number of messages sent within QQ groups in the past 30 days.
\end{itemize}

We have 298,837,578 friend relationships among these users.  1,773 groups are labeled as ``game'' and the remaining groups are labeled as ``non-game''.

We construct the hierarchical graph from this Tencent QQ group data as follows.  A user is treated as an object, and a QQ group as a graph instance.  The users in one group are connected by their friendship.  The attribute matrix $X$ is filled with the attribute values of the users.  The statistics of the graph instances are listed in Table \ref{tab:sqq}.  We build the hierarchical graph from the graph instances via common members across groups.  That is, if groups $A$ and $B$ have more than one common member, we connect them.

\begin{table}
  \caption{Statistics of collected Tencent QQ groups}
  \label{tab:sqq}
  \begin{tabular}{ccccc}
    \toprule
    \textbf{Class label}&\textbf{Number}&\textbf{Nodes}&\textbf{Edges}&\textbf{Density} \\
    \midrule
	game&1,773&147&395&5.48\%\\
	non-game&36,063&365&1586&3.28\%\\
  \bottomrule
\end{tabular}

\raggedright{The node and edge numbers and density are the average for each type of QQ group.}
\eat{\vspace{-0.3cm}}
\end{table}

\subsubsection{Baselines and Metrics}
We use the same set of baselines as in Section \ref{syn.base}.  1000 graph instances are used as labeled training instances for all methods except SEAL-AI and active-SAGE-Cheby, for which only 500 are used as labeled training instances at hand and then $B$ is set to 500 for active learning.  We use 10,000 instances for testing for all methods.  We run each method 3 times and report its average accuracy.  The hyperparameters of SAGE are the same as the settings in Section \ref{bench.base}. Since the class distribution is quite imbalanced in this data set, we report the Macro-F1 instead of accuracy.

\subsubsection{Results}

\begin{figure}
\begin{center}
\includegraphics [width=0.45\textwidth,scale=1]{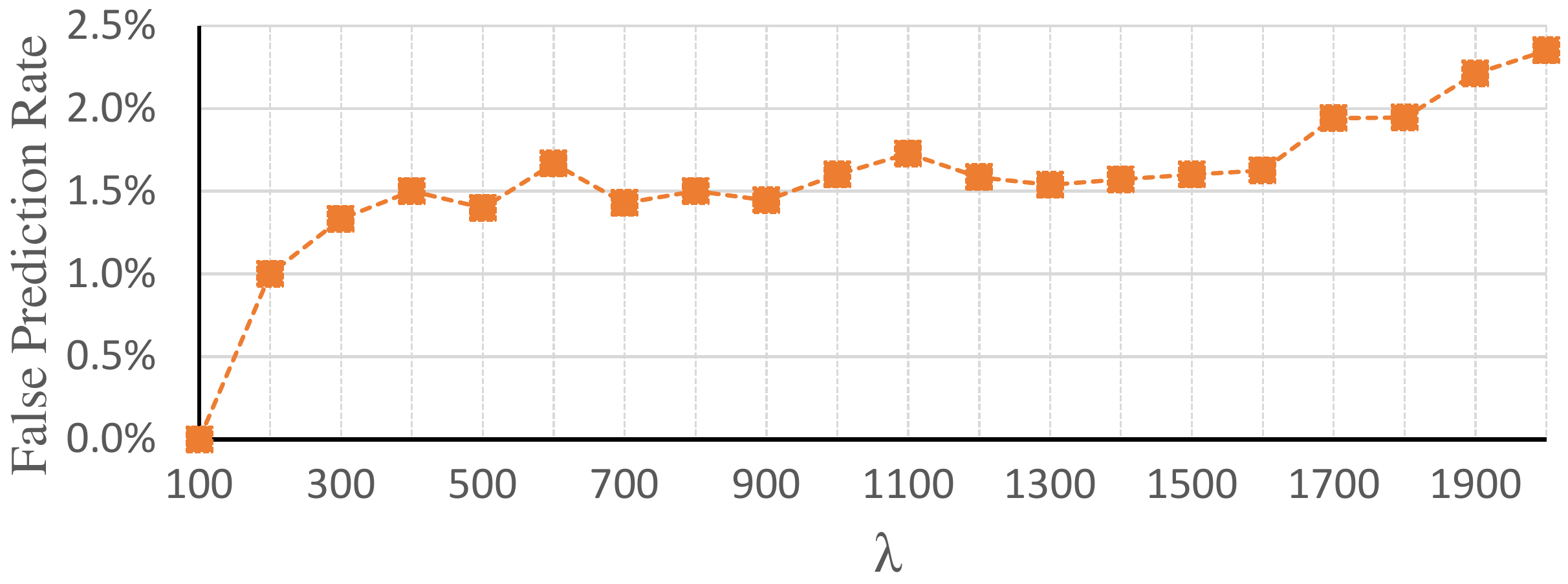}
\end{center}

\caption{The false prediction rate of GCN with $\lambda$ in SEAL-CI.}
\label{fig.seal}
\eat{\vspace{-0.3cm}}
\end{figure}

Table \ref{tab:qq} shows the experimental results.  SEAL-C/AI outperform GK, WL and grah2vec by at least 12\% in Macro-F1.  Within our framework, GCN is better than Cheby-GCN for about 6\%.  SEAL-AI outperforms SEAL-CI by 2.4\%.  Next we provide the reason why SEAL-AI outperforms SEAL-CI on this data set.  Figure \ref{fig.seal} shows the false prediction rate (i.e., the percentage of misclassified instances) within the $\lambda$ most confident predictions of GCN.  As we can see, the false prediction rate increases as $\lambda$ increases and it reaches $2.4\%$ when $\lambda=2000$. In the framework of SEAL-CI, as the iteration goes on, we shall bring in more noise to the parameter update of SAGE, while all the training samples in SEAL-AI are informative and correct. This explains why SEAL-AI outperforms SEAL-CI on this Tencent QQ group data.

 \eat{\begin{table}
  \caption{Features of users}
  \vspace{-0.3cm}
  \label{tab:general}
  \scalebox{0.7}{
  \begin{tabular}{lll}
    \toprule
    \textbf{Type} & \textbf{Notation} & \textbf{Meaning} \\
    \midrule
	\multirow{4}{*}{Profile information}&Registered days& M=male, F=female\\
	&Province ID&65 years and older\\
	&City ID&years of education\\
	&No. of friends&body mass index\\
	\hline
	\multirow{2}{*}{Behavioral features}&No. of online&medical history of diabetes: Y=Yes, N=No\\
	&Online days&medical history of heart disease: Y=Yes, N=No\\
	\hline
	\multirow{2}{*}{Physical test}&GRIPAM& maximum grip strength (kg)\\
	&GAITSPEED&walking speed (m/s)\\
  \bottomrule
\end{tabular}
}
\vspace{-0.3cm}
\end{table}}
\begin{table}
  \caption{Comparison of different methods on Tencent QQ group data for semi-supervised graph classification}
  \label{tab:qq}
  \scalebox{0.9}{
  \begin{tabular}{ccc}
    \toprule
     &\textbf{Algorithm}&\textbf{Macro-F1} \\
    \midrule
		\multirow{2}{*}{*1}& \textbf{GK-SVM} & 48.8\%\\
		& \textbf{WL-SVM} & 47.8\%\\
		\hline
		\multirow{1}{*}{*2} & \textbf{graph2vec-GCN} &48.1\%\\
		\hline
		\multirow{2}{*}{*3}& \textbf{cautious-SAGE-Cheby} & 64.3\%\\
		& \textbf{active-SAGE-Cheby} & 66.7\%\\
		\hline
		\multirow{3}{*}{*4} & \textbf{SAGE} & 54.7\% \\
		& \textbf{SEAL-CI} & 70.8\%\\
		& \textbf{SEAL-AI} & \textbf{73.2}\%\\
	  \bottomrule
\end{tabular}
}
\end{table}
\subsubsection{Visualization}
\begin{figure}
\begin{center}
\includegraphics [width=0.45\textwidth,scale=1]{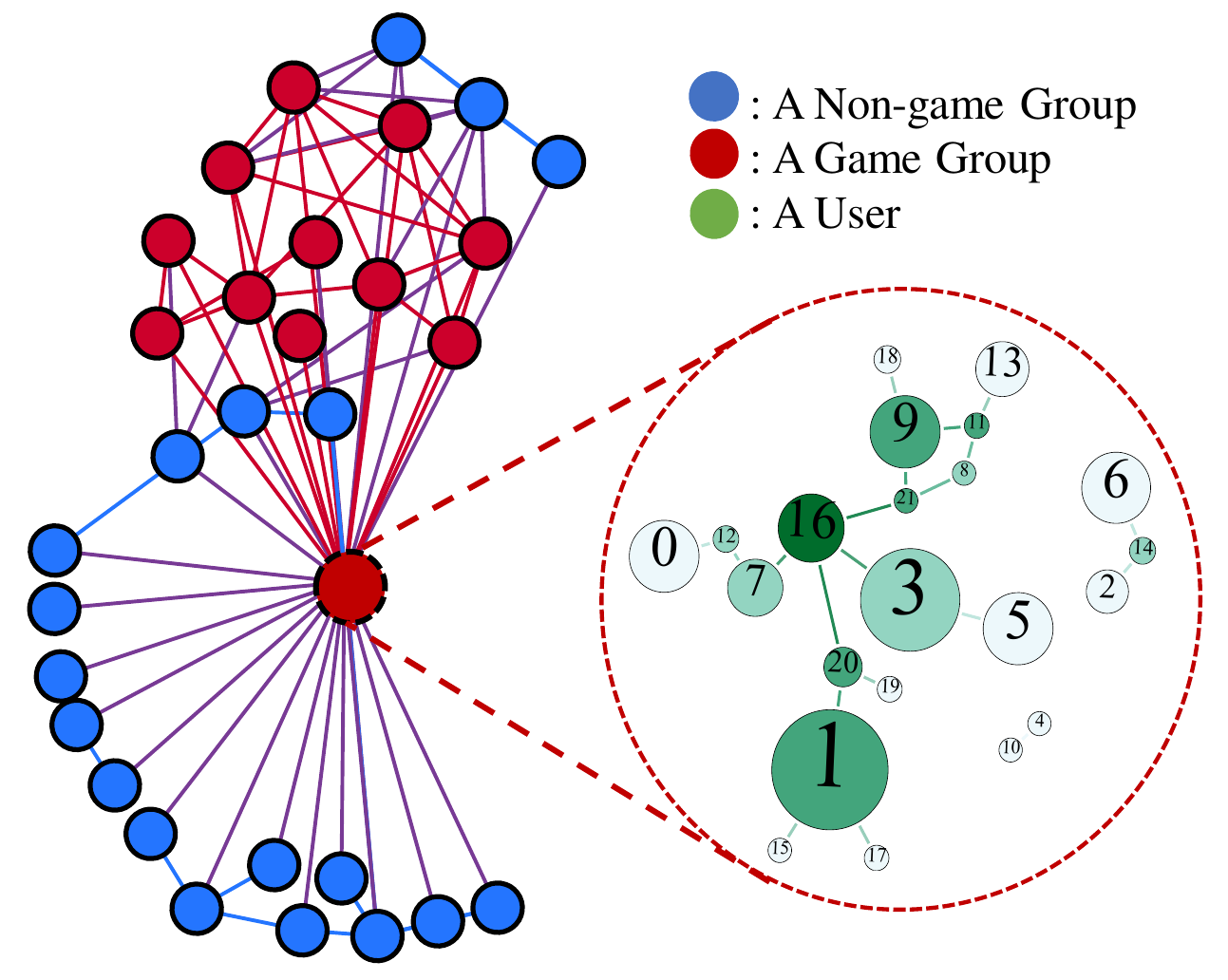}
\end{center}

\caption{The ego network of a ``game'' group.  The left side is the ego network, in which ``game'' groups are in red and ``non-game'' groups are in blue.  The right side is the internal structure of the ego ``game'' group, in which a bigger node indicates a larger importance, and a darker color implies a larger node degree.}
\label{fig.gam}
\vspace{-0.3cm}
\end{figure}

We provide visualization of a ``game'' group and its neighborhood in Figure \ref{fig.gam}.  The left part is the ego network of the center ``game'' group.  In the one-hop neighborhood of this ``game'' group, there are 10 ``game'' groups and 19 ``non-game'' groups. ``Game'' groups are densely interconnected with a density of 34.5\%, whereas ``non-game'' groups are sparsely connected with a density of 8.8\%.  The much higher density among ``game'' groups validates that common membership is an effective way to relate them in a hierarchical graph for classification.  The right part depicts the internal structure of the ego ``game'' group with 22 users. A bigger node indicates a larger importance, and a darker green color implies a larger node degree.  These 22 members are loosely connected and there are no triangles.  This makes sense because in reality online ``game'' groups are not acquaintance networks.  Regarding the learned node importance, node 1 has the highest importance as it is the second active member and has a large degree in this group.  Node 16 is also important since it has the highest degree in this group.  The ``border'' member 5 has a big attention weight since it has the largest number of days ever since the registration day and is quite active in this group.
\subsubsection{Discussion}
How to construct a hierarchical graph from raw data is an open question.  In the above experiment, we connect two QQ groups if they have more than one common member (i.e., $>1$).  When we change the threshold, it directly affects the edge density in the hierarchical graph, and may influence the classification performance.  For example, if we connect two QQ groups when they have one common member or more (i.e., $\geq 1$), the edge density is 2.8\% compared with 0.27\% in the first setting.  A proper setting of this threshold is data dependent, and can be determined through a validation set.

\section{Related Work}\label{sec.related}
This work is related to semi-supervised classification of networked data, variable-sized graph embedding and active learning.

Most work on semi-supervised learning for networked data aims to utilize the network structure to boost the learning performance.  The assumption is that network context can provide additional information that is not covered by node attributes.  Ever since the pioneer work of Sen et al.\ \cite{sen2008collective}, Iterative Classification Algorithm (ICA) has become a paradigm for networked data with limited annotations.  In ICA, for each node a local classifier takes the estimated labels of its neighborhood and its own features as input, and outputs a new estimated label.  The iteration continues until adjacent estimations stabilize.  In ALFNET ~\cite{bilgic2010active}, the authors first cluster the network nodes into several groups, and design a content-only classifier CO and a collective classifier CC.  Based on the disagreement score of CO and CC in each iteration, a candidate instance set is generated from different clusters and labeled.  Then both CO and CC are re-trained using the labeled set until convergence.  One main difference between ICA and ALFNET is that ICA does not require human intervention while ALFNET needs human annotation in case labels of the candidate set are not available. \eat{This is an active learning method and it is worthwhile here since the number of samples to be labeled is small compared with the total number of samples.}

Recent work has focused on using deep learning neural networks to further improve the performance. ~\cite{yang2016revisiting} leverages both network context and node features by jointly training node embedding to predict the class label and the context of the network.  Later Kipf and Welling~\cite{kipf2017semi} simplify the loss design by only considering the supervised loss while network context is exploited by the GCN operator.  Our problem setting is different from all of the above, as the node is no longer a fixed-size feature vector but a variable-size graph.  It can be regarded as a generalization of the previous setting, and cannot be handled by existing solutions effectively.  %However since the previous work can not take a hierarchical graph as input, we need a new approach to tackle this challenge.

Representation learning on graphs has been proposed to transform instances in topological space into fixed-size vectors in Euclidean space in which geometric distance reflects their structural similarity.  There are two trends on this topic, one of which is a shift from node embedding~\cite{perozzi2014deepwalk,grover2016node2vec} to whole graph embedding.  \cite{Yanardag:2015:DGK:2783258.2783417} uses CBOW and skip-gram model~\cite{mikolov2013distributed}, previously proven to be successful in natural language processing, to learn a new graph kernel.  Meanwhile, some other methods focus on generating graph embeddings by integrating node embeddings.  \cite{Niepert2016LearningCN} proposes a spatial-based graph CNN operator and then concatenates these obtained node representations by imposing a problem-specific node ordering.  \cite{defferrard2016convolutional} defines a ``graph coarsening'' operation by first clustering the node representations and then applying a max-pooling operation.  However, all these methods need some preprocessing steps such as node ordering or clustering, which is not a necessity from a data-driven perspective.   Another trend is a shift from unsupervised embedding~\cite{mikolov2013distributed} to supervised embedding~\cite{dai2016discriminative,DBLP:journals/corr/LinFSYXZB17}, which provides better performance for downstream classification tasks.  In this sense, our embedding method SAGE performs whole graph embedding in a supervised way.

Active learning has been integrated in many collective classification methods \cite{settles2012active,bilgic2010active} to find the most informative samples to be labeled.  However, research that generalizes active learning with deep semi-supervised learning is still lacking.  The closest work is \cite{zhou2017fine} in which the authors utilize active learning to incrementally fine-tune a CNN network for image classification.  Our solution SEAL-AI is different in the sense that the informative samples selected by active learning are used to update the parameters of the graph embedding network, whose output is then fed into HC in an iterative framework.

\section{CONCLUSION}\label{sec.con}

In this paper, we study semi-supervised graph classification from a hierarchical graph perspective.  The hierarchical graph is a much too complicated input for classification, thus we first design a supervised, self-attentive graph embedding method SAGE to embed graph instances into fixed-length vectors, which are a common input form for classification.  We build two classifiers IC and HC at the graph instance level and the hierarchical graph level respectively to fully exploit the available information.  Our semi-supervised solutions SEAL-C/AI adopt an iterative framework to update IC and HC alternately with an enlarged training set.  Experimental results on synthetic graphs and Tencent QQ group data show that SEAL-C/AI outperform other competitors by a significant margin in accuracy/Macro-F1, and they also generate meaningful interpretations of the learned representations for graph instances.

\begin{acks}
The authors would like to thank Tencent Security Platform Department for discussions and suggestions. The work described in this paper was supported by grants from the Research Grant Council of the Hong Kong Special Administrative Region, China [Project No.: CUHK 14205618], Tencent AI Lab Rhino-Bird Focused Research Program GF201801 and the CUHK Stanley Ho Big Data Decision Analytics Research Centre.

\end{acks}

\bibliographystyle{ACM-Reference-Format}
\balance
\bibliography{sample-bibliography}

\end{document}